\documentclass[lettersize,journal]{IEEEtran}
\usepackage{amsmath,amsfonts}
\usepackage{algorithmic}
\usepackage{algorithm}
\usepackage{array}
\usepackage[caption=false,font=normalsize,labelfont=sf,textfont=sf]{subfig}
\usepackage{textcomp}
\usepackage{stfloats}
\usepackage{url}
\usepackage{verbatim}
\usepackage{graphicx}
\usepackage{cite}
\usepackage{booktabs}
\usepackage{multirow}
\newcommand\defeq{\mathrel{\stackrel{\makebox[0pt]{\mbox{\normalfont\scriptsize def}}}{:=\,}}}

\usepackage{tikz}
\newcommand\copyrighttext{%
  \footnotesize This work has been submitted to the IEEE for possible publication. Copyright may be transferred without notice, after which this version may no longer be accessible.}
\newcommand\copyrightnotice{%
\begin{tikzpicture}[remember picture,overlay]
\node[anchor=south,yshift=10pt] at (current page.south) {\fbox{\parbox{\dimexpr\textwidth-\fboxsep-\fboxrule\relax}{\copyrighttext}}};
\end{tikzpicture}%
}

\begin{document}

\title{Safety-Aligned 3D Object Detection: Single-Vehicle, Cooperative, and End-to-End Perspectives}

\author{Brian Hsuan-Cheng Liao, Chih-Hong Cheng, Hasan Esen, Alois Knoll
\thanks{Brian Hsuan-Cheng Liao and Hasan Esen are with DENSO AUTOMOTIVE Deutschland GmbH, 85386 Eching, Germany. {\tt\small {h.liao, h.esen}@eu.denso.com}}%
\thanks{Chih-Hong Cheng is with the Department of Computer Science, Carl von Ossietzky University of Oldenburg, 26129 Oldenburg, Germany. {\tt\small chih-hong.cheng@uni-oldenburg.de }}%
\thanks{Brian Hsuan-Cheng Liao and Alois Knoll are with the Department of Computer Engineering, Technical University of Munich, 85748 Garching, Germany. {\tt\small k@tum.de}}%
}

\maketitle
\copyrightnotice

\begin{abstract}

Perception plays a central role in connected and autonomous vehicles (CAVs), underpinning not only conventional modular driving stacks, but also cooperative perception systems and recent end-to-end driving models. While deep learning has greatly improved perception performance, its statistical nature makes perfect predictions difficult to attain. Meanwhile, standard training objectives and evaluation benchmarks treat all perception errors equally, even though only a subset is safety-critical. In this paper, we investigate safety-aligned evaluation and optimization for 3D object detection that explicitly characterize high-impact errors. Building on our previously proposed safety-oriented metric, NDS-USC, and safety-aware loss function, EC-IoU, we make three contributions. First, we present an expanded study of single-vehicle 3D object detection models across diverse neural network architectures and sensing modalities, showing that gains under standard metrics such as mAP and NDS may not translate to safety-oriented criteria represented by NDS-USC. With EC-IoU, we reaffirm the benefit of safety-aware fine-tuning for improving safety-critical detection performance. Second, we conduct an ego-centric, safety-oriented evaluation of AV--infrastructure cooperative object detection models, underscoring its superiority over vehicle-only models and demonstrating a safety impact analysis that illustrates the potential contribution of cooperative models to “Vision Zero.” Third, we integrate EC-IoU into SparseDrive and show that safety-aware perception hardening can reduce collision rate by nearly $30\%$ and improve system-level safety directly in an end-to-end perception-to-planning framework. Overall, our results indicate that safety-aligned perception evaluation and optimization offer a practical path toward enhancing CAV safety across single-vehicle, cooperative, and end-to-end autonomy settings.

\end{abstract}


\section{Introduction}

\IEEEPARstart{C}{onnected} and autonomous vehicles (CAVs) have achieved major milestones over the past decades, including public demonstrations, pilot programs, and commercial deployments~\cite{ponyai2025autonomous,waymo2025worlds}. Nevertheless, incidents involving CAVs continue to occur~\cite{nhtsa2025order}, underscoring that safe and reliable autonomous driving (AD) remains an open and pressing challenge.

We address this challenge from the perception layer of the AD stack. Perception forms the information boundary between high-dimensional sensor measurements and the structured world model consumed by downstream prediction, planning, and control modules. Consequently, perception errors can propagate through the stack and are often difficult to compensate for the downstream: a planner cannot reliably avoid obstacles that are missed, mis-localized, or whose spatial extent is underestimated without becoming overly conservative and degrading performance. Although modern perception systems are strongly powered by deep learning (DL), fundamental limitations still constrain their reliability. For instance, DL models can be vulnerable to small input perturbations, i.e., adversarial examples~\cite{evtimov2018robust}. 

\begin{figure}
    \centering
    \hspace{0.03\linewidth}
    \includegraphics[width=0.95\linewidth]{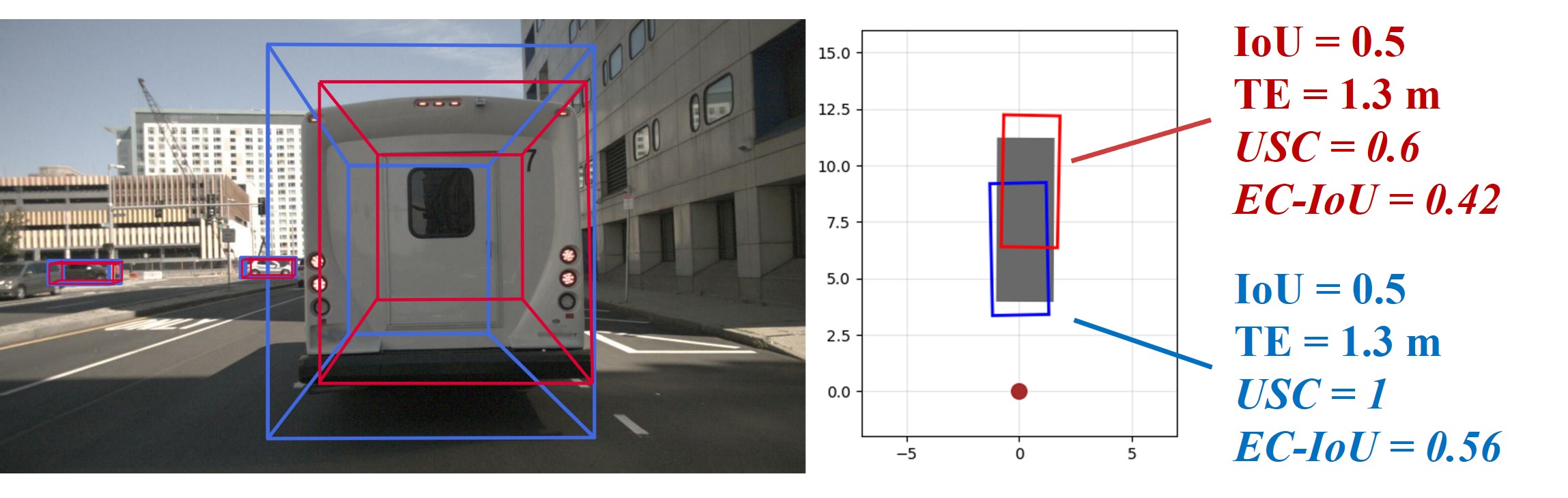}
    \caption{Asymmetry in the safety consequences of perception errors: The red prediction overestimates the truck’s longitudinal distance and under-covers it relative to the blue prediction, making it more safety-critical. Whereas standard metrics such as IoU and translation error (TE) assign identical values to both predictions, our safety-aligned metrics---USC and EC-IoU---favor the blue prediction as the safer one.}
    \label{fig:safety_critical_errors}
\end{figure}

In this work, we focus on a pervasive limitation---the statistical nature of DL---which inevitably yields predictions that deviate from ground truths, potentially inducing safety risk during driving. Nonetheless, one key observation is that \emph{perception errors are not equally consequential}: some directly create hazardous situations, while others are relatively benign. This asymmetry motivates safety-aligned perception: \emph{rather than optimizing average performance, a perception model should explicitly address errors that are most consequential to safety}. Our prior work operationalized this principle in two ways: (i) Uncompromising Spatial Constraints (USC), a safety-oriented metric that evaluates whether predictions conservatively cover ground truth from the ego perspective and, when integrated into Nuscenes Detection Score (NDS), shows strong correlation with collision rate in closed-loop simulation~\cite{liao2024usc}; and (ii) a safety-aware fine-tuning strategy that reweights optimization toward safety-critical regions of the ground truth via ego-centric intersection-over-union (EC-IoU)~\cite{liao2024eciou}. Figure~\ref{fig:safety_critical_errors} illustrates this intuition with two imperfect predictions for a truck ahead of the ego vehicle. Although both predictions are inaccurate, the red one fails to cover the truck’s ego-facing extent, which may lead downstream modules to underestimate occupancy and thus increase collision risk. This article substantially extends the line of research in three directions.

First, we present an expanded study of 3D object detection models on nuScenes~\cite{caesar2020nuscenes}, covering a broad range of nine models with different architectures and sensing modalities. The results show that improvements under standard metrics such as mAP and NDS do not necessarily translate into gains under safety-oriented criteria, highlighting the latter as important indicators for model development and selection. We further show that safety-aware fine-tuning consistently improves safety-critical detection performance and standard accuracy.

Second, we extend safety-oriented evaluation from single-vehicle perception to AV--infrastructure cooperative perception. Using the TUMTraf benchmark~\cite{zimmer2024tumtrafv2x}, we evaluate cooperative object detectors from the ego-vehicle perspective with safety-oriented matching. The results confirm the superiority of cooperative models, but also reveal an important limitation: they can introduce localization biases that raise concerns for ego-vehicle safety due to perspective changes. While such biases may be mitigated by the proposed safety-aware fine-tuning strategy, we focus on extending the evaluation study to a safety impact analysis that illustrates the potential contribution of cooperative perception toward the ``Vision Zero'' goal at intersection scenarios~\cite{johansson2009vision}.

Third, we show that safety-aware perception hardening also benefits system-level safety in the emerging end-to-end optimization paradigm. By integrating our EC-IoU loss into SparseDrive~\cite{sun2024sparsedrive}, a state-of-the-art perception-to-planning model, we reduce collision rate by nearly 30\%. In summary, the main contributions of this paper are:
\begin{itemize}
    \item An expanded study of 3D object detectors across diverse architectures and sensing modalities, showing that gains on conventional benchmarks may not reflect improvements under safety-oriented criteria, and that safety-aware fine-tuning consistently strengthens general and safety-critical performance.
    \item An safety-oriented, ego-centric evaluation protocol for AV--infrastructure cooperative perception, together with an intersection-level impact analysis toward the ``Vision Zero'' objective.
    \item A demonstration that safety-aware perception hardening transfers to end-to-end models, directly reducing the collision rate and improving system-level safety.
\end{itemize}

\section{Related Work}
\label{sec:related_work}

\subsection{Safety-Oriented Evaluation}
\label{sec:rw:safety_eval}

Safety-oriented evaluation of perception aims to bridge the gap between generic accuracy metrics and safety-relevant downstream performance. Standard object detection protocols typically rely on overlap- or distance-based true-positive (TP) measures such as Intersection-over-Union (IoU) in the KITTI benchmark~\cite{geiger2012are} and translation error (TE) as in the nuScenes benchmark~\cite{caesar2020nuscenes}. While effective for ranking average localization fidelity, these measures treat many error modes similarly and may not reflect the safety-criticality of specific mislocalizations, e.g., under-covering the ego-facing extent of a nearby obstacle.

A growing line of work therefore designs task-specific or safety-oriented metrics. For instance, planner-centric evaluation directly measures how perception affects planning, e.g., by comparing planner outputs induced by predicted objects versus ground truth~\cite{philion2020learning}. Complementary to planner-centric approaches, several works introduce safety-motivated geometric criteria. Deng \emph{et al.} proposed Support Distance Error (SDE), evaluating longitudinal and lateral support distances between predictions and ground truths relative to the ego vehicle~\cite{deng2021revisiting}. Mori \emph{et al.} explored safety-oriented adaptations of evaluation protocols and pass/fail criteria inspired by human perception performance~\cite{mori2024safety}. In contrast to approaches that rely on planner access or specialized object representations, our USC metric adopts simple constraints defined on common 3D bounding boxes and is applicable across sensing modalities, while being validated against system-level collision rate in closed-loop simulation.

\subsection{Safety-Aware Fine-Tuning}
\label{sec:rw:safety_finetune}

Improving object detection performance via training objectives has been studied extensively. Beyond classification-focused losses (e.g., focal loss) and standard regression losses (e.g., $\ell_1/\ell_2$), many works propose overlap-aware objectives to better align predicted boxes with ground truth, including the standard IoU loss~\cite{yu2016unitbox} and its derivations~\cite{rezatofighi2018generalized,zhang2022focal,he2022alphaiou}. Other approaches incorporate uncertainty estimation or probabilistic modeling to better capture depth ambiguity and calibration, especially for monocular 3D object detection~\cite{wang2021probabilistic}.

Safety-aware fine-tuning has also been explored through re-weighting strategies that emphasize critical classes (e.g., vulnerable road users) or difficult samples~\cite{cheng2020safety,lyssenko2024safety}. Our work follows a similar strategy but focuses on ego-centric safety-critical localization rather than class importance alone. Specifically, we proposed EC-IoU as a graded, ego-centric objective that emphasizes safety-critical regions of ground truths while respecting full accuracy.

\subsection{Cooperative Perception}
\label{sec:rw:coop}

Cooperative perception leverages V2X communication to fuse observations from neighboring vehicles and/or roadside infrastructure, improving perception under occlusions, limited sensor range, and adverse conditions. Methods are commonly categorized into \emph{early}, \emph{intermediate}, and \emph{late} fusion, depending on whether raw measurements, learned features, or object hypotheses are exchanged~\cite{huang2025v2xsurvey}. The field has been accelerated by strong models and extensive benchmarks such as OPV2V~\cite{xu2022opv2v}, V2X-ViT~\cite{xu2022v2xvit}, DAIR-V2X~\cite{yu2022dairv2x}, and TUMTraf-V2X~\cite{zimmer2024tumtrafv2x}. 

While most prior work primarily targets higher detection accuracy under global or infrastructure-centric evaluation protocols, we evaluate the models from the ego vehicle's viewpoint using our safety-oriented criteria. In doing so, we examine whether cooperation perception reduces safety-critical mislocalizations that most directly affect the ego vehicle’s decision making, e.g., ego-facing under-coverage or longitudinal overestimation of nearby obstacles. This perspective complements existing benchmarking practice~\cite{yu2022dairv2x,zimmer2024tumtrafv2x} and naturally enables a safety impact analysis of cooperative perception at intersections.

\subsection{End-to-End Driving}
\label{sec:rw:e2e}

End-to-end (E2E) driving aims to map sensor inputs directly to planning or control outputs using a unified model. Early E2E driving models often predicted a single trajectory or control sequence~\cite{pomerleau1988alvinn,bojarski2016end2end}, while recent methods increasingly adopt structured intermediate representations (e.g., detections, tracks, and maps) and conduct joint optimization across different modules such as perception, prediction, and planning to improve interpretability and performance. UniAD is a representative framework that unifies perception, prediction, and planning via multiple Transformers and intermediate supervision~\cite{hu2023uniad}. More recently, SparseDrive proposed a sparse scene representation and a parallel prediction--planning design to improve efficiency and planning safety~\cite{sun2024sparsedrive}.

Moreover, vision-language-action (VLA) models extend E2E driving by incorporating language understanding and reasoning capabilities, enabling action explanations and human interactivity~\cite{jiang2025vla4ad}. Current VLA research explores how to closely couple vision-language backbones with action prediction, how to ground reasoning in traffic context, and how to ensure safety and reliability under open-world conditions. While promising, such work often requires substantial data and compute resources. In this work, we investigate enhancing the safety of E2E driving models by strengthening the perception component in a lightweight manner. Specifically, we integrate our safety-aware EC-IoU loss into SparseDrive training, demonstrating that lightweight, safety-oriented perception hardening can also improve system-level safety under joint optimization.

\section{Foundation: USC and EC-IoU}
\label{sec:foundation}

This section presents the safety alignment measures used throughout the paper: (i) \emph{Uncompromising Spatial Constraints (USC)} and the derived metric \emph{NDS-USC} for safety-oriented evaluation~\cite{liao2024usc}, and (ii) \emph{Ego-Centric IoU (EC-IoU)} as a more granular safety-aware overlap measure for both evaluation and fine-tuning~\cite{liao2024eciou}.

\subsection{USC: Uncompromising Spatial Constraints}
\label{sec:foundation:usc}

Consider a matched pair of a predicted 3D bounding box $P$ and its ground-truth box $G$. USC encodes two safety-driven localization requirements that are particularly relevant to collision avoidance: (1) the prediction should enclose the ground truth in the perspective view (PV), and (2) the prediction should not be farther than the ground truth in the bird's-eye view (BEV) along the ego-facing side.

\subsubsection{PV enclosure and IoGT}
\label{sec:foundation:usc:pv}

\begin{figure}[t]
    \centering
    \includegraphics[width=0.98\linewidth]{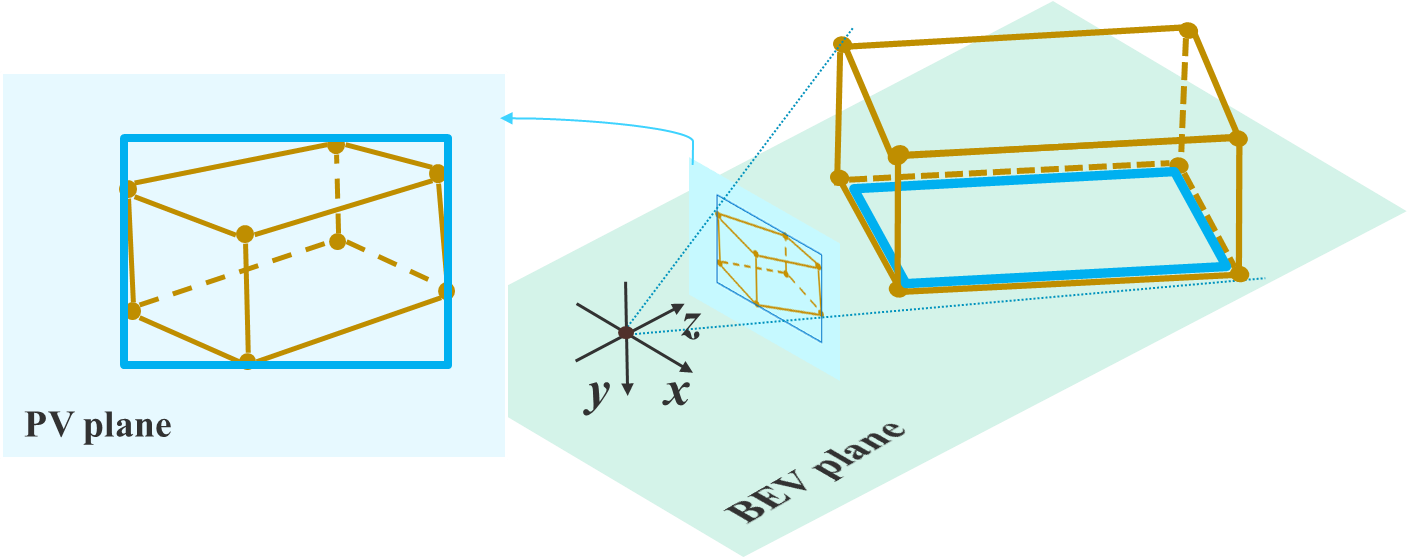}
    \caption{USC applies perspective-view (PV) and bird's-eye-view (BEV) projections to enforce two safety-relevant constraints. Adapted from Fig.~2 in~\cite{liao2024usc}~\textcopyright~2024~IEEE.}
    \label{fig:projections}
\end{figure}

As Fig.~\ref{fig:projections} illustrates, USC uses perspective projection to map the 3D boxes to the PV plane. For a 3D point $p \defeq (x_p,y_p,z_p)$, the PV coordinates $(a,b)$ are obtained by the standard pinhole projection:
\begin{equation}
\label{eq:pv_projection}
\begin{bmatrix}
a \\ b \\ f
\end{bmatrix}
=
\frac{f}{z_p}
\begin{bmatrix}
x_p \\ y_p \\ z_p
\end{bmatrix}.
\end{equation}
Projecting all eight box corners and taking the min/max coordinates yields axis-aligned PV boxes $P^{PV}$ and $G^{PV}$. The PV constraint requires enclosure:
\begin{equation}
\label{eq:pv_constraint}
\Pi^{PV} \defeq \big(G^{PV} \subset P^{PV}\big).
\end{equation}
To quantify the enclosure degree, USC adopts Intersection-over-Ground-Truth (IoGT):
\begin{equation}
\label{eq:iogt}
\mathrm{IoGT}(P^{PV},G^{PV}) \defeq \frac{\mathrm{Area}(P^{PV}\cap G^{PV})}{\mathrm{Area}(G^{PV})},
\end{equation}
which saturates at $1$ when $P^{PV}$ fully encloses $G^{PV}$.

\subsubsection{BEV distance underestimation and ADR}
\label{sec:foundation:usc:bev}

PV enclosure alone may still allow a prediction that is farther but large enough to enclose $G^{PV}$. USC therefore adds a BEV constraint that ensure distance underestimation on the closest point and ego-visible extreme corners, so that no region of the ground truth is exposed to the ego perspective. Using orthographic projection, we obtain BEV rectangles $P^{BEV}$ and $G^{BEV}$. Let $v_G^c$ and $v_P^c$ be the closest points to the ego vehicle (assumed at origin), and let $v_G^l,v_G^r$ and $v_P^l,v_P^r$ be the ego-visible left/right extreme corners of $G^{BEV}$ and $P^{BEV}$, respectively. We first constrain the closest point with
\begin{equation}
\label{eq:bev_closest}
\Pi^{BEV}_c \defeq \big(\lVert v_P^c\rVert \le \lVert v_G^c\rVert\big),
\end{equation}
and then enforce consistency of the ego-visible extremes via
\begin{equation}
\label{eq:bev_constraint}
\Pi^{BEV} \defeq \Pi^{BEV}_c \wedge \neg \perp\big(\{v_P^c v_P^r, v_P^c v_P^l, v_G^c v_G^r, v_G^c v_G^l\}\big),
\end{equation}
where $\perp(\cdot)$ indicates intersection of the specified line segments (excluding overlapping endpoints).

For a quantitative BEV score, USC defines an Average Distance Ratio (ADR) using the three representative points $i\in\{c,r,l\}$: 
\begin{equation}
\label{eq:adr}
\mathrm{ADR}(P^{BEV},G^{BEV}) \defeq
\left(
\prod_{i\in\{c,r,l\}}
\max\!\left(1,\frac{\lVert v_G^i\rVert}{\lVert v_P^i\rVert}\right)
\right)^{\!\!1/3}.
\end{equation}
ADR saturates at $1$ when all three representative points of $P^{BEV}$ are no farther than their ground-truth counterparts.

\subsubsection{USC score and NDS-USC}
\label{sec:foundation:usc:consolidation}

USC consolidates the PV and BEV constraints into a qualitative predicate 
\begin{equation}
\label{eq:usc_predicate}
\Pi^{USC} \defeq \Pi^{PV} \wedge \Pi^{BEV},
\end{equation}
and defines a true-positive score 
\begin{equation}
\label{eq:usc_score}
\mathrm{USC} \defeq \mathrm{IoGT}\times \mathrm{ADR}.
\end{equation}
Similar to mean Average Precision (mAP), averaging USC over matched pairs of an object class yields $\mathrm{AUSC}$; averaging over classes yields $\mathrm{mAUSC}$. Finally, to incorporate false positives and false negatives, USC augments the NuScenes Detection Score (NDS) to form the overall safety-oriented metric:
\begin{equation}
\label{eq:nds_usc}
\mathrm{NDS\!-\!USC} \defeq \frac{1}{2}\big(\mathrm{NDS}+\mathrm{mAUSC}\big).
\end{equation}

\subsubsection{Closed-Loop Validation}
\label{sec:foundation:usc:validation}

To validate the system-safety relevance of USC, our prior study compared model-level perception metrics with the collision rate observed in closed-loop simulation~\cite{liao2024usc}. Tab.~\ref{tab:usc_correlation} shows the absolute Pearson correlation coefficients between each metric and the simulated vehicle collision rate. USC-based metrics exhibit stronger correlation than conventional detection metrics, with NDS-USC achieving the highest correlation. This result supports USC and NDS-USC as more safety-relevant evaluation measures.

\begin{table}[t]
\caption{Absolute Pearson correlation coefficients between model-level metrics and simulated system-level collision rate. Higher values indicate stronger correlation with collision risk.}
\label{tab:usc_correlation}
\centering
\begin{tabular}{lc}
\toprule
Metric & Correlation \\
\midrule
mAP     & 0.699 \\
NDS     & 0.806 \\
mAUSC   & 0.814 \\
NDS-USC & \textbf{0.925} \\
\bottomrule
\end{tabular}
\end{table}

\subsection{EC-IoU: Ego-Centric Intersection-over-Union}
\label{sec:foundation:eciou}

Although USC correlates strongly with system-level collision rate, it can saturate once its safety constraints are satisfied and thus may no longer distinguish between predictions. To provide a more graded safety-aware measure, EC-IoU assigns higher importance to the safety-critical parts of the ground truth, namely those closer to the ego vehicle.

\subsubsection{Ego-centric weighting and weighted area}
\label{sec:foundation:eciou:weight}

On the BEV plane, let $P$ and $G$ denote the predicted and ground-truth oriented 2D polygons. Let $\rho(x,y)$ be the Euclidean distance from $(x,y)$ to the ego vehicle at the origin, and let $(x_G,y_G)$ be the center of $G$. EC-IoU defines the weighting function over $G$ as
\begin{equation}
\label{eq:eciou_weight}
\omega_G(x,y) \defeq \left[\frac{\rho(x_G,y_G)}{\rho(x,y)}\right]^{\alpha},\qquad \alpha\ge 0,\ (x,y)\in G.
\end{equation}
Accordingly, the weighted area of a region $D\subseteq G$ is defined as
\begin{equation}
\label{eq:weighted_area}
\mathrm{Weighted\mbox{-}Area}_G(D) \defeq \iint_{D} \omega_G(x,y)\,dA.
\end{equation}

\subsubsection{EC-IoU definition}
\label{sec:foundation:eciou:def}

Using $\mathrm{Weighted\mbox{-}Area}_G(\cdot)$, EC-IoU is defined as
\begin{equation}
\label{eq:eciou}
\begin{split}
&\mathrm{EC\mbox{-}IoU}(P,G)\\
&\defeq\frac{\mathrm{Weighted\mbox{-}Area}_G(P\cap G)}
{\mathrm{Weighted\mbox{-}Area}_G(G)+\mathrm{Area}(P)-\mathrm{Area}(P\cap G)}.
\end{split}
\end{equation}
This formulation preserves the boundedness of IoU while placing greater emphasis on coverage of the ego-near, and thus more safety-critical, portion of $G$.

\subsubsection{Efficient approximation of weighted areas}
\label{sec:foundation:eciou:approx}

Computing $\mathrm{Weighted\mbox{-}Area}_G(\cdot)$ in closed form is generally intractable. Inspired by the Mean Value Theorem, we approximate the weighted area of a convex polygon $D$ as the product of its area and an estimated mean weight:
\begin{equation}
\label{eq:mvt_weighted_area}
\mathrm{Weighted\mbox{-}Area}_G(D) \approx \omega_{G,D}\cdot \mathrm{Area}(D).
\end{equation}
For a polygon $D$ with vertices $\{(x_i^D,y_i^D)\}_{i=1}^{m}$, the mean weight is estimated by the geometric mean of the vertex weights:
\begin{equation}
\label{eq:mean_weight_geom}
\omega_{G,D} \defeq \left(\prod_{i=1}^{m}\omega_G(x_i^D,y_i^D)\right)^{1/m}.
\end{equation}
This approximation yields EC-IoU values close to numerical integration while remaining efficient for training and evaluation. 

\begin{figure}[]
\centering
    \includegraphics[width=0.65\linewidth]{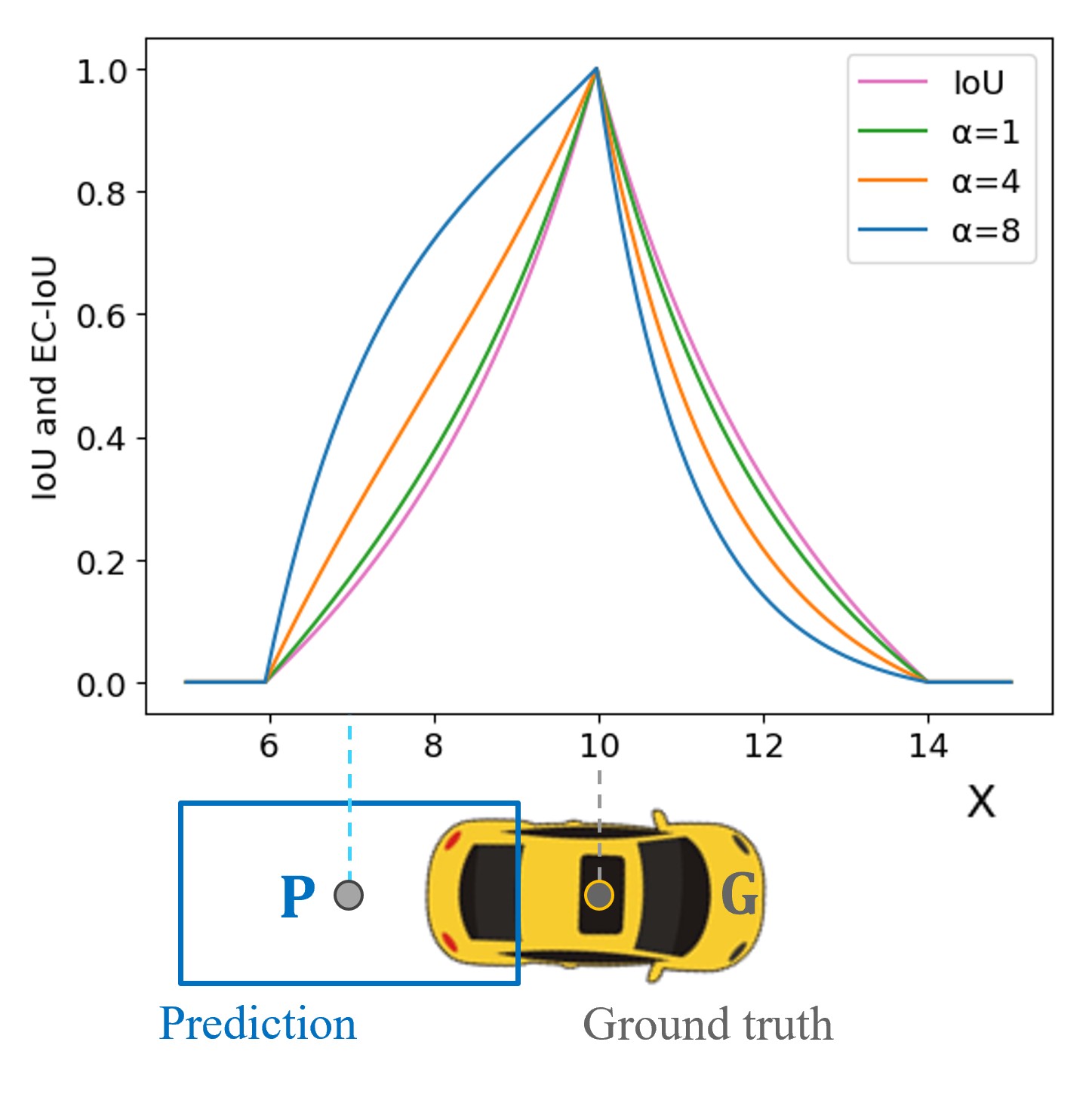}
    \caption{IoU and EC-IoU evaluation (with different $\alpha$ values) as a prediction moves over the ground truth along the x-axis. The ego vehicle is assumed at $x=0$, the ground truth $G$ is centered at $x=10$, and the blue box denotes an example prediction $P$ centered at $x=7$. Adapted from Fig.~3 in~\cite{liao2024eciou}~\textcopyright~2024~IEEE.}
    \label{fig:ec_iou_valuation}
\end{figure}

Fig.~\ref{fig:ec_iou_valuation} compares the effect of different levels of ego-centric weighting on EC-IoU against standard IoU. In terms of complexity, EC-IoU remains comparable to ordinary IoU. For each prediction--ground-truth pair, IoU computation requires finding the vertices of the intersection polygon, sorting them, and computing the resulting area. The additional weighting step in EC-IoU only requires evaluating the weights of the intersection vertices and thus introduces linear overhead in the number of vertices. Since two oriented rectangles have at most eight intersection vertices and each weight evaluation takes constant time, the extra time cost is negligible in practice.

\section{Expanded Study on Single-Vehicle Perception}
\label{sec:single_vehicle}

This section presents an expanded study of safety-oriented evaluation and safety-aware fine-tuning for 3D object detection from a single vehicle. We first benchmark representative camera-, lidar-, and fusion-based object detectors on nuScenes using conventional and safety-oriented metrics. We then demonstrate the benefit of safety-aware fine-tuning by fine-tuning the sub-optimal models.

\subsection{Safety-Oriented Benchmarking}
\label{sec:single_vehicle:benchmark}

\subsubsection{Experimental Setup}
\label{sec:single_vehicle:evaluation:setup}

We use the nuScenes benchmark~\cite{caesar2020nuscenes}. In addition to standard metrics (i.e., mAP and NDS), we report the safety-oriented error measure mAUSC and the integrated safety-oriented metric NDS-USC.

We benchmark representative state-of-the-art 3D object detectors from three categories: (i) camera-based models, including FCOS3D~\cite{wang2021fcos3d}, PGD~\cite{wang2021probabilistic}, DETR3D~\cite{wang2021detr3d}, and PETR~\cite{liu2022petr}; (ii) lidar-based models, including PointPillars~\cite{lang2019pointpillars}, SSN~\cite{zhu2020ssn}, and CenterPoint~\cite{yin2021centerpoint}; and (iii) camera--lidar fusion, represented by BEVFusion~\cite{liu2023bevfusion}, which is a fusion framework utilizing bird's-eye-view features and also supports a lidar-only configuration. For consistency, all models are evaluated using strong public checkpoints from the unified MMDetection3D platform~\cite{mmdet3d2020}. Tab.~\ref{tab:models} briefly summarizes the key contributions of each model.

\begin{table}[t]
\caption{Representative models included in our benchmark and their key contributions.}
\label{tab:models}
\centering
\footnotesize
\setlength{\tabcolsep}{4pt}
\begin{tabular}{lp{0.67\linewidth}}
\toprule
\textbf{Model} & \textbf{Key Contributions} \\
\midrule
FCOS3D~\cite{wang2021fcos3d} & Projects 3D coordinates onto 2D feature maps and adapts a fully convolutional one-stage detector for 3D box regression. \\

PGD~\cite{wang2021probabilistic} & Extends FCOS3D with probabilistic depth estimation and geometric relation graphs, enabling uncertainty-aware depth refinement for more accurate localization. \\

DETR3D~\cite{wang2021detr3d} & Adopts a Transformer architecture in which sparse 3D queries retrieve encoded 2D image features and decode them directly into 3D bounding boxes. \\

PETR~\cite{liu2022petr} & Improves DETR3D by injecting 3D coordinate information into 2D image features, yielding 3D position-aware feature maps for more accurate localization. \\

PointPillars~\cite{lang2019pointpillars} & Converts point clouds into pseudo-images using vertical pillars, enabling efficient processing by CNN backbones. \\

SSN~\cite{zhu2020ssn} & Extends PointPillars by incorporating explicit shape information from point clouds through an additional shape-aware loss. \\

CenterPoint~\cite{yin2021centerpoint} & Represents each object by its center point in bird's-eye view using a keypoint heatmap, and regresses 3D dimensions and orientation from the detected centers. \\

BEVFusion~\cite{liu2023bevfusion} & Extends CenterPoint by fusing camera and lidar inputs in a shared bird's-eye-view feature space for joint 3D detection. \\
\bottomrule
\end{tabular}
\end{table}

\begin{figure}[]
    \centering
    \includegraphics[width=0.98\linewidth]{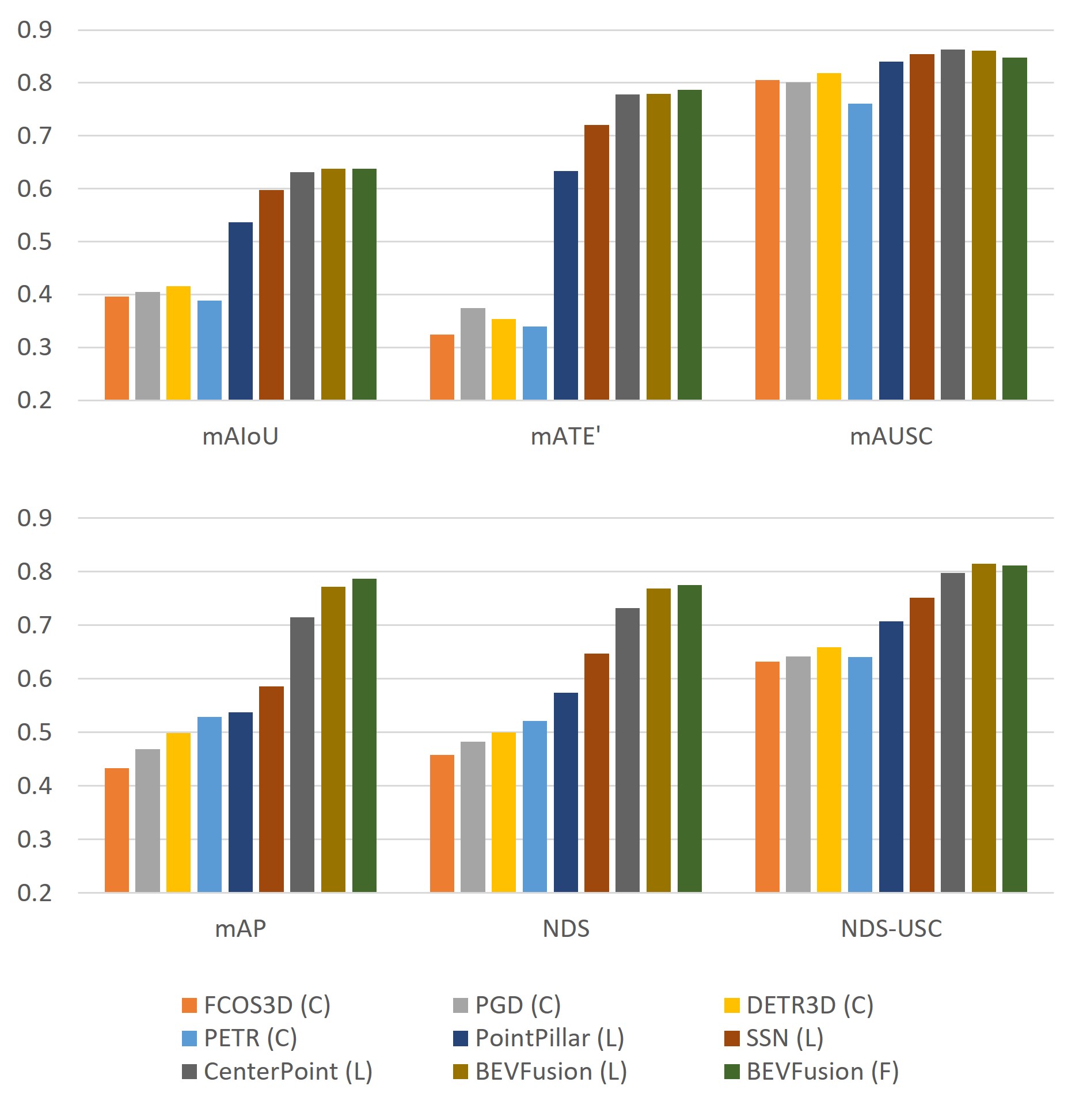}
    \caption{Benchmarking results of representative 3D object detectors on nuScenes, labeled by the sensor modality (C: camera, L: lidar, and F: fusion). \textit{Top:} true-positive error measures (mAIoU, mATE$'$, mAUSC). \textit{Bottom:} overall metrics (mAP, NDS, NDS-USC). Safety-oriented metrics tolerate certain non-critical errors but emphasize safety-critical mislocalizations, changing the performance picture compared with standard metrics.}
    \label{fig:single:benchmark}
\end{figure}

\subsubsection{Results and Discussion}
\label{sec:single_vehicle:evaluation:result}

Fig.~\ref{fig:single:benchmark} summarizes benchmarking results at two levels. The top plot compares true-positive error measures in mAIoU, mATE$'\defeq 1-\text{mATE}$, and mAUSC, while the bottom plot compares overall performance using mAP, NDS, and NDS-USC.

Three observations are noted. First, conventional error measures mAIoU and mATE$'$ strongly penalize camera-based detectors, compared to mAUSC. This is due to USC’s design: conservative box oversizing and distance underestimation are less safety-critical and should not be strongly penalized. Second, safety-oriented evaluation can identify suboptimal models that are less obvious under standard metrics. For example, PGD and BEVFusion remain strong under conventional metrics, yet exhibit weaker safety-oriented detection performance under mAUSC, suggesting non-negligible safety-critical error modes. Third, these trends propagate to overall metrics: while mAP and NDS reflect familiar ranking trends from the literature, NDS-USC shows that apparent gains may diminish when safety-critical detection performance is emphasized. Collectively, these results motivate safety-oriented evaluation as a complementary lens for model selection in safety-critical autonomy stacks.

\subsection{Safety-Aware Fine-Tuning}
\label{sec:single_vehicle:finetune}

\subsubsection{Experimental Setup}
\label{sec:single_vehicle:finetune:setup}

We now examine safety-aware fine-tuning using EC-IoU. We focus on the camera-based detectors PGD and PETR, which are the most accurate CNN- and Transformer-based models in our benchmark, respectively, yet both are suboptimal under safety-oriented evaluation with mAUSC and NDS-USC.

For EC-IoU, we use $\alpha\in\{2,4\}$ during training and set $\alpha=2$ for true-positive error measuring. For each model, we first train a baseline checkpoint for 12 epochs on nuScenes, and then continue fine-tuning for 6 additional epochs using either the original regression loss or the EC-IoU loss. All experiments are conducted with a batch size of 32 on a server with four NVIDIA L40S GPUs. Fine-tuning takes about 16 hours for PETR and 18 hours for PGD. For each model, we report the average over three baseline runs and three EC-IoU fine-tuning runs.

\subsubsection{Results and Discussion}
\label{sec:single_vehicle:finetune:result}

Tab.~\ref{tab:single:eciou_finetune} shows that EC-IoU fine-tuning consistently improves both conventional and safety-oriented measures for PGD and PETR. For PGD, EC-IoU variants contribute to score improvement differently, with EC-IoU-2 yielding the best IoU, EC-IoU, TE$'$, and NDS, while EC-IoU-4 gives the best USC, mAP, and NDS-USC. For PETR, the gains are consistent: EC-IoU-4 achieves the best performance on all reported metrics. Overall, the results indicate that EC-IoU can not only improve safety-critical detection performance but also enhance standard standard accuracy.

\begin{figure}[]
    \centering
    \includegraphics[width=0.8\linewidth]{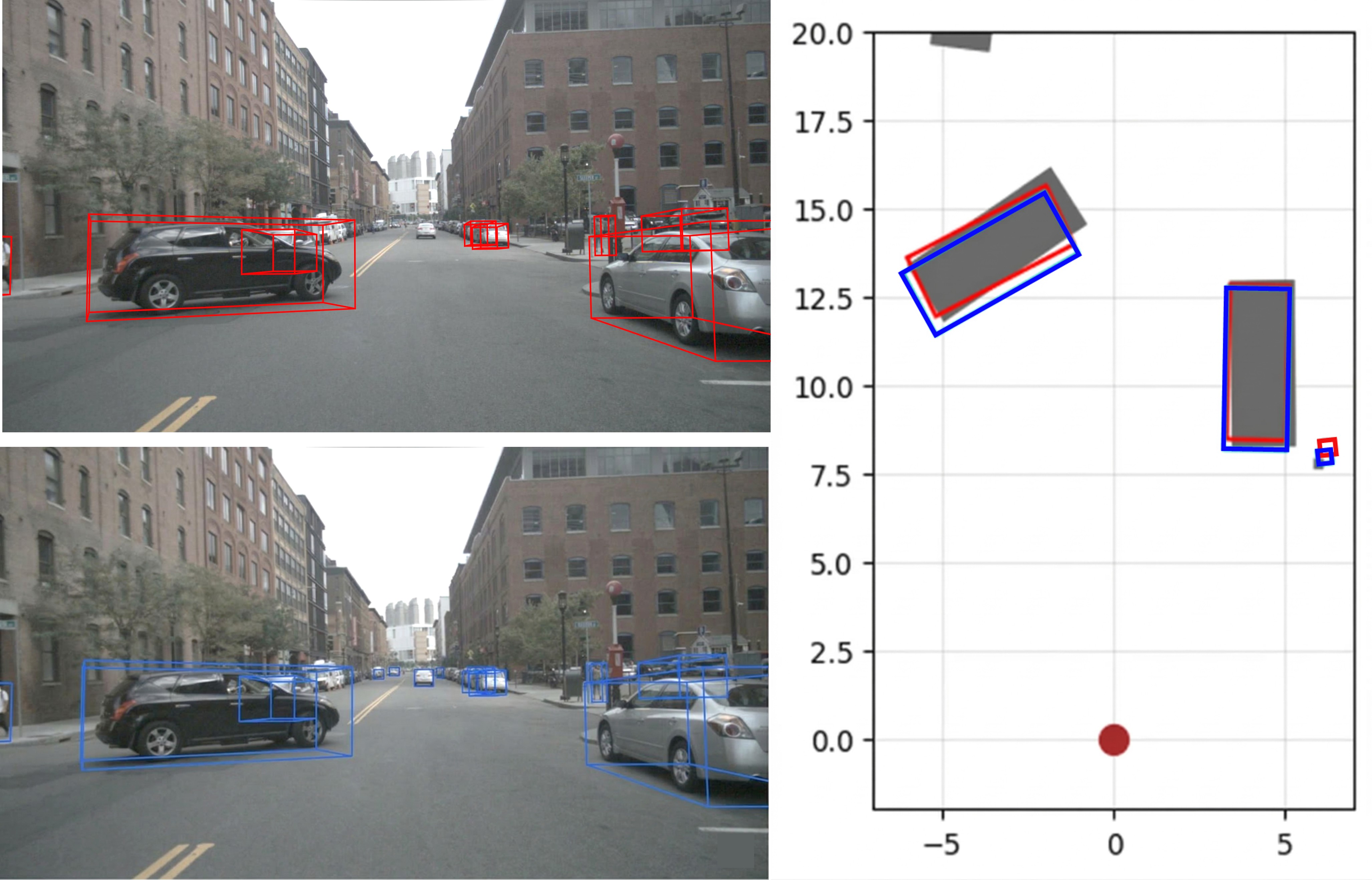}
    \caption{Qualitative comparison of PETR fine-tuned with the standard loss (red) and our safety-aware EC-IoU loss (blue). EC-IoU shifts predictions toward the ego vehicle, improving coverage of safety-critical regions.}
    \vspace{-2mm}
    \label{fig:single:nuscenes_eciou}
\end{figure}

Fig.~\ref{fig:single:nuscenes_eciou} illustrates the mechanism behind these gains. Compared with the baseline prediction, EC-IoU shifts the box closer to the ego vehicle and improves coverage of the safety-critical ego-facing regions. This behavior is consistent with the design of EC-IoU, which assigns higher importance to ground-truth regions nearer to the ego vehicle. 

\begin{table}[]
\caption{Fine-tuning PGD and PETR with standard and EC-IoU losses (higher is better).}
\label{tab:single:eciou_finetune}
\centering
\footnotesize
\setlength{\tabcolsep}{3pt}
\begin{tabular}{lccccccc}
\toprule
\multirow{2}{*}{Model} &
\multicolumn{4}{c}{TP measures $\uparrow$} &
\multicolumn{3}{c}{Overall (\%) $\uparrow$} \\
\cmidrule(lr){2-5}\cmidrule(lr){6-8}
& IoU & EC-IoU & TE$'$ & USC & mAP & NDS & \scriptsize{NDS-USC} \\
\midrule
PGD          & 0.404 & 0.397 & 0.374 & 0.801 & 46.86 & 48.29 & 64.19 \\
w/ EC-IoU-2  & \textbf{0.419} & \textbf{0.408} & \textbf{0.408} & \underline{0.817} & \underline{47.34} & \textbf{48.87} & \underline{65.28} \\
w/ EC-IoU-4  & \underline{0.417} & \underline{0.405} & \underline{0.381} & \textbf{0.819} & \textbf{47.62} & \underline{48.82} & \textbf{65.36} \\
\midrule
PETR         & 0.389 & 0.360 & 0.340 & 0.761 & 52.83 & 52.07 & 64.10 \\
w/ EC-IoU-2  & \underline{0.393} & \underline{0.363} & \underline{0.347} & \underline{0.770} & \underline{53.25} & \underline{52.40} & \underline{64.68} \\
w/ EC-IoU-4  & \textbf{0.395} & \textbf{0.365} & \textbf{0.349} & \textbf{0.771} & \textbf{53.46} & \textbf{52.52} & \textbf{64.79} \\
\bottomrule
\end{tabular}
\end{table}

\vspace{-2mm}
\subsection{Takeaways}
\label{sec:single_vehicle:takeaways}

\begin{itemize}
    \item \textbf{Safety-oriented evaluation reveals complementary model behavior:} Gains under mAP and NDS do not necessarily translate to gains under NDS-USC, and some high-accuracy models still exhibit safety-critical localization weaknesses.
    \item \textbf{Safety-aware fine-tuning delivers general improvement:} Our EC-IoU loss not only enhances safety-critical localization but also increases classical accuracy terms.
\end{itemize}

\section{Ego-Centric Safety-Oriented Evaluation of Cooperative Perception and Its Safety Impact}
\label{sec:coop}

In this section, we extend safety-oriented evaluation from single-vehicle perception to AV--infrastructure cooperative perception and study its potential safety impact. By augmenting the ego vehicle's sensing with roadside observations, cooperative perception can enhance visibility and mitigate occlusions in dense traffic scenarios such as intersections~\cite{zimmer2024tumtrafv2x}. Prior studies mainly report intersection-wide detection performance from a roadside viewpoint. Here, we ask a more ego-safety-oriented question: \emph{Does cooperation improve perception for objects that matter to the automated vehicle when performance is evaluated from the ego viewpoint and with a safety-oriented criterion?} To answer this question, we adopt the TUMTraf benchmark and adapt its evaluation protocol accordingly.

\subsection{TUMTraf Benchmark and CoopDet3D}
\label{sec:coop:benchmark}

\begin{figure}[]
    \centering
    \includegraphics[width=0.98\linewidth]{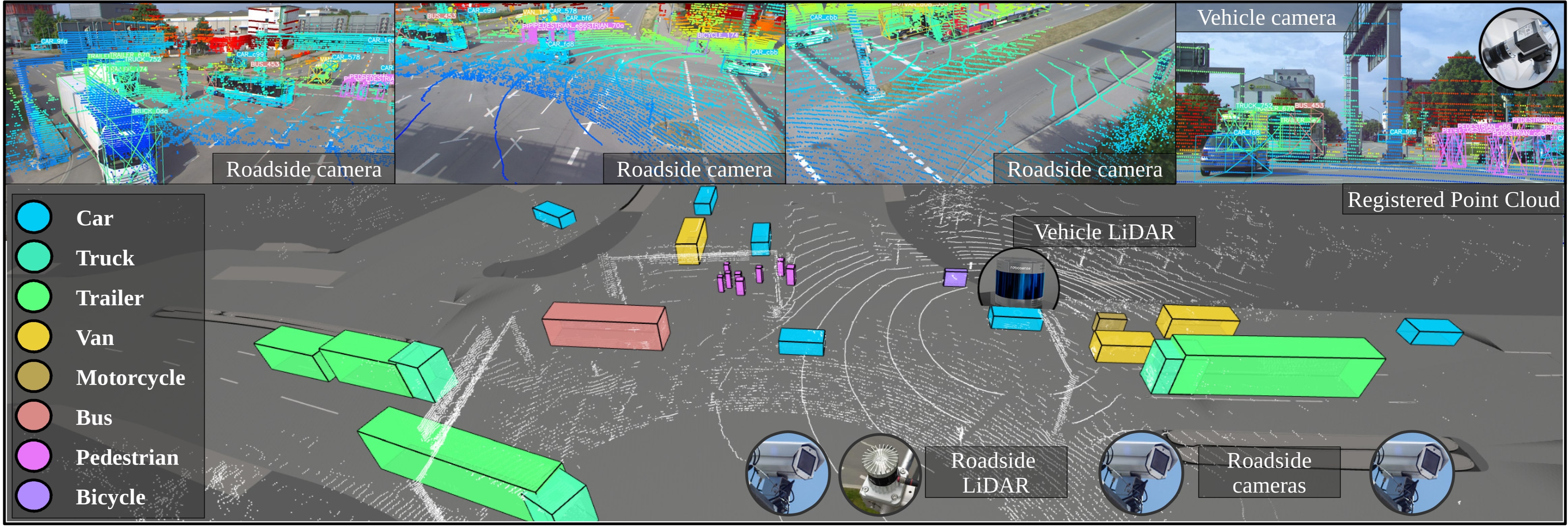}
    \caption{TUMTraf cooperative perception benchmark~\cite{zimmer2024tumtrafv2x}. \textit{Top:} three roadside camera views and the ego-vehicle camera view. \textit{Bottom:} Overview of the intersection covered by vehicle and infrastructure sensors. Adapted from~\cite{zimmer2024tumtrafv2x} \textcopyright~2024 IEEE.}
    \label{fig:coop:tumtraf_overview}
\end{figure}

Our study is based on TUMTraf, a large benchmark for AV--infrastructure cooperative 3D object detection~\cite{zimmer2024tumtrafv2x}. As shown in Fig.~\ref{fig:coop:tumtraf_overview}, TUMTraf provides synchronized sensing from an ego vehicle with a front-facing camera and a 360\textdegree~lidar, together with roadside cameras and a 360\textdegree~lidar, covering an intersection near TUM in Garching, Germany. The benchmark also provides a reference model, CoopDet3D~\cite{zimmer2024tumtrafv2x}, which extracts BEV features separately for the vehicle and infrastructure using a BEVFusion-style backbone~\cite{liu2023bevfusion} and fuses them by feature-level max pooling. CoopDet3D supports nine configurations formed by three sensing modalities---camera, lidar, and fusion---and three domains---vehicle, infrastructure, and cooperation.

\subsection{Ego-Centric Safety-Oriented Evaluation Protocol}
\label{sec:coop:protocol}

TUMTraf follows a KITTI-style evaluation protocol~\cite{geiger2012are} and reports results from a roadside camera viewpoint using IoU-based matching with a low universal true-positive threshold, i.e., $\tau_c = 0.1$ for any class $c$~\cite{zimmer2024tumtrafv2x}. While suitable for general intersection coverage, this protocol does not directly reflect ego-safety relevance. We therefore make three adaptations.

First, we distinguish roadside-view and ego-view evaluation. Roadside-view follows the original benchmark and considers all labeled objects visible from a selected roadside camera. Ego-view restricts evaluation to objects relevant to the ego vehicle by discarding objects outside the ego camera field of view or fully occluded in the ego view. This yields roadside-view mAP and ego-view mAP, denoted by RV-mAP and EV-mAP.

Second, we replace the original universal matching threshold with KITTI-style class-dependent thresholds:
\[
\tau_{\mathrm{car\text{-}like}}=0.7,\quad
\tau_{\mathrm{bicycle\text{-}like}}=0.5,\quad
\tau_{\mathrm{pedestrian}}=0.3.
\]
We also exclude empty cases when averaging AP. This avoids systemic underestimation caused by assigning zero AP to cases without ground-truth instances.

Third, to emphasize safety-aware localization, we replace IoU with EC-IoU as the affinity measure for matching predictions to ground truths. This yields a ego-centric safety-oriented metric, denoted EC-mAP, in which a prediction is counted as a true positive only if its EC-IoU exceeds the class-dependent matching threshold. In effect, EC-IoU penalizes predictions placed farther than the ground truth from the ego vehicle while tolerating closer, more conservative placement, thereby emphasizing ego-safety-critical localization errors.

\begin{table}[]
\caption{CoopDet3D performance ($\%$) under roadside-view, ego-view, and ego-centric safety-oriented evaluation on TUMTraf~\cite{zimmer2024tumtrafv2x}.}
\label{tab:coop:all}
\footnotesize
\centering
\setlength{\tabcolsep}{4pt}
\begin{tabular}{llccc}
\toprule
Domain & Mod. & RV-mAP & EV-mAP & EC-mAP \\
\midrule
Vehicle & Cam.   & 16.85 & 28.56 & 30.98 \\
Vehicle & lidar  & 25.12 & 56.98 & 55.09 \\
Vehicle & Fusion & 34.90 & 61.92 & 60.94 \\
Coop.   & Cam.   & 33.00 & 54.11 & 43.33 \\
Coop.   & lidar  & 38.53 & 67.49 & 55.76 \\
Coop.   & Fusion & 46.35 & 71.11 & 68.44 \\
\bottomrule
\end{tabular}
\end{table}

\begin{figure}[]
    \centering
    \includegraphics[width=0.9\linewidth]{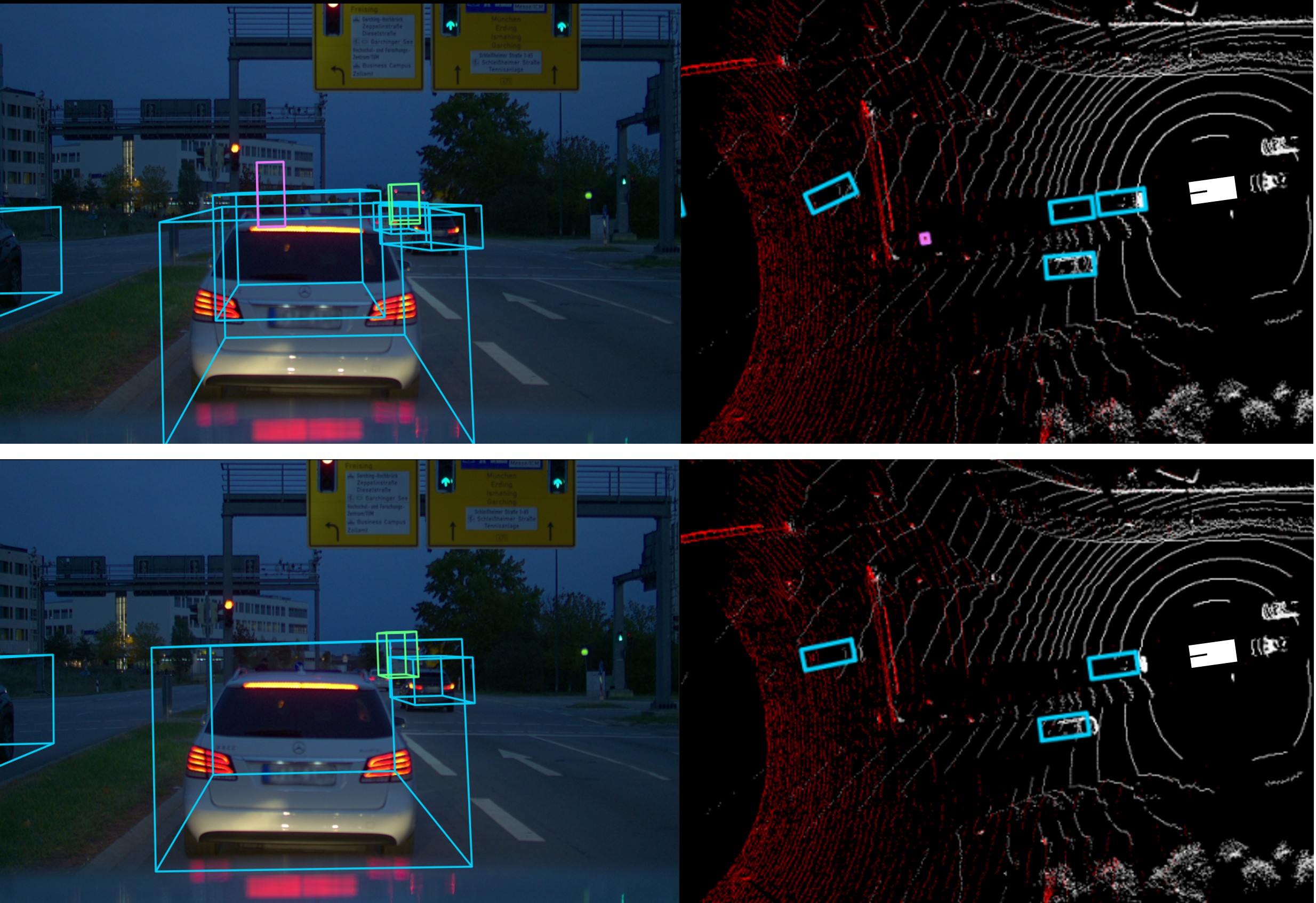}
    \caption{\textit{Top:} vehicle-only lidar-based model. \textit{Bottom:} cooperative lidar-based model. The cooperative model reduces false positives, but slightly overestimates the lead vehicle’s longitudinal distance from the ego perspective. In the bird’s-eye view, the roadside lidar point cloud is shown in red, while the ego vehicle appears on the right in white and is moving leftward.}
    \vspace{-2mm}
    \label{fig:coop:coop_misalignment}
\end{figure}

\vspace{-2mm}
\subsection{Results and Discussion}
\label{sec:coop:results}

Tab.~\ref{tab:coop:all} compares CoopDet3D under roadside-view, ego-view, and ego-centric safety-oriented evaluation. Moving from roadside-view to ego-view increases the scores of all configurations, mainly because the evaluation scope changes from all objects visible to a roadside camera to those relevant and visible to the ego vehicle. Importantly, cooperative configurations retain a clear advantage over vehicle-only configurations under ego-view evaluation across all sensing modalities.

Comparing EV-mAP and EC-mAP reveals a different pattern. Most configurations incur a drop under EC-mAP, indicating that predictions are often placed slightly farther than the ground truth from the ego viewpoint, which is safety-critical under our metric definition. Notably, this drop is more pronounced for the cooperative configurations, although they all remain superior to the vehicle-only counterparts. This suggests that perception cooperation may introduce small localization biases, i.e., a ``box-pulling'' effect. Such an effect cannot be captured by standard IoU-based evaluation alone.

Fig.~\ref{fig:coop:coop_misalignment} illustrates this effect: the cooperative lidar-based model suppresses several false positives but slightly misaligns the prediction of the lead vehicle in an ego-critical manner. Overall, the results answer our research question positively: \emph{cooperative perception improves ego-relevant perception even under ego-view and safety-oriented evaluation. Nevertheless, it may introduce small detrimental biases near the ego vehicle}. This reveals a concrete optimization target---reducing ego-centric localization biases in the near field---which can be addressed through safety-aware fine-tuning with EC-IoU loss. We leave it as important future work.

\vspace{-2mm}
\subsection{Safety Impact Analysis}
\label{sec:coop:impact}

We turn our focus to how much cooperative perception may reduce collisions at an intersection. To obtain an estimate, we consider three stages of traffic composition: human-driven vehicles only, mixed traffic with AVs, and mixed traffic with AVs supported by cooperative perception. Tab.~\ref{tab:coop:safety} summarizes the estimated annual collisions for the three stages, and the following provides the rationale.

\begin{table}[]
\caption{Estimated annual collisions at an intersection with human-driven vehicles (HDVs) only, mixed traffic with HDVs and AVs, and mixed traffic with AVs supported by cooperative perception (CP).}
\label{tab:coop:safety}
\centering
\footnotesize
\setlength{\tabcolsep}{5pt}
\begin{tabular}{lc}
\toprule
\textbf{Stage} & \textbf{Collisions/year} \\
\midrule
HDVs only      & 10.95 \\
HDVs + AVs     & 5.48 \\
HDVs + AVs + CP & \textbf{2.67} \\
\bottomrule
\end{tabular}
\end{table}

For the HDV-only stage, we use a historical average collision rate of $1.5$ collisions per million entering vehicles at signalized intersections~\cite{fhwa2013signalized} together with a representative traffic volume of $20{,}000$ vehicles/day~\cite{bast2025automatische}, yielding $10.95$ collisions/year. For the mixed HDVs+AVs stage, we refer to a recent commercial safety report~\cite{dilillo2024autonomous} and assume a conservative AV collision reduction ratio of $50\%$, resulting in $5.48$ collisions/year.

For the HDVs+AVs+CP stage, rather than assuming another fixed reduction ratio, we relate collision reduction to the improvement in safety-oriented perception performance, here EC-mAP, using a linear regression model:
\begin{equation}
\Delta Col \defeq \beta \cdot \Delta EC\text{-}mAP,
\label{eq:coop:linear}
\end{equation}
with
\begin{equation}
\beta \defeq \rho \cdot \frac{\sigma_{Col}}{\sigma_{EC\text{-}mAP}},
\label{eq:coop:slope}
\end{equation}
where $\rho$ denotes the correlation coefficient between perception performance and collision rate, and $\sigma_{Col}$ and $\sigma_{EC\text{-}mAP}$ are the corresponding standard deviations~\cite{montgomery2021intro}. For perception performance, we set $\sigma_{EC\text{-}mAP}=5$ based on typical benchmark variability~\cite{dong2023benchmarking}. For collision rate, we model it as a Poisson variable, giving $\sigma_{Col}\approx\sqrt{\mu}$ with $\mu=5.48$ from the second stage~\cite{kalra2016driving}. Finally, we take $\rho=-0.8$ as a conservative setting guided by the mAUSC and NDS-USC results in Tab.~\ref{tab:usc_correlation} and $\Delta EC\text{-}mAP=7.5$, corresponding to the improvement from the vehicle-only fusion model to the cooperative fusion model in Tab.~\ref{tab:coop:all}. This yields $\Delta Col=2.81$ and a final estimate of $2.67$ collisions/year, corresponding to an additional reduction of approximately $51.3\%$.

To further probe this result, we vary the AV collision reduction ratio in the second stage. If AVs reduce collisions by only $30\%$, the residual collision rate becomes $7.67$ collisions/year in the second stage and $3.32$ in the third stage, corresponding to a further $56.7\%$ reduction due to cooperative perception. If AVs reduce collisions by $70\%$, the residual collision rate becomes $3.29$ and $1.11$ collisions/year in the second and third stages, respectively, corresponding to a further $66.2\%$ reduction. In both cases, cooperative perception plays an important role in driving the collision rate lower. Under the same linear model, if AVs reduce collisions by $86.8\%$, leaving $1.44$ collisions/year, cooperative perception could in principle eliminate the remaining collisions. In this sense, AVs and cooperative perception may jointly approach the ``Vision Zero'' goal~\cite{johansson2009vision}.

It is noted that this analysis relies on strong assumptions, including the AV collision reduction ratio, the linear mapping from EC-mAP to collision reduction, and the chosen parameter settings. Nonetheless, it provides an initial estimate of how cooperative perception may reduce residual intersection collisions. This also motivates more rigorous future evaluation in closed-loop simulation and long-term field studies.

\vspace{-2mm}
\subsection{Takeaways}
\label{sec:coop:takeaways}

\begin{itemize}
    \item \textbf{Cooperative perception improves ego-safety-relevant localization:} Cooperative models consistently outperform vehicle-only baselines across sensing modalities, even under ego-centric safety-oriented evaluation using EC-mAP. However, EC-mAP also exposes subtle misalignment tendencies by cooperative models, thereby identifying a concrete target for future safety-aware optimization.
    \item \textbf{Cooperative perception may eliminate residual collisions toward ``Vision Zero":} Our safety impact analysis suggests that, at sufficient AV penetration, cooperative perception can play an important role in driving the intersection collision rate toward zero.
\end{itemize}

\section{Safety-Aware Perception Fine-Tuning for End-to-End Driving}
\label{sec:e2e}

We next investigate whether safety-aware perception fine-tuning remains beneficial under the emerging end-to-end (E2E) driving paradigm. Unlike modular stacks, which expose detection outputs to a separate planner and optimize the modules independently, modern E2E driving models jointly optimize intermediate perception tasks such as object detection and tracking together with downstream planning. We hypothesize that injecting a safety-aware signal into the perception component can improve \emph{system-level safety} through joint optimization.

\begin{figure}[]
    \centering
    \includegraphics[width=0.98\linewidth]{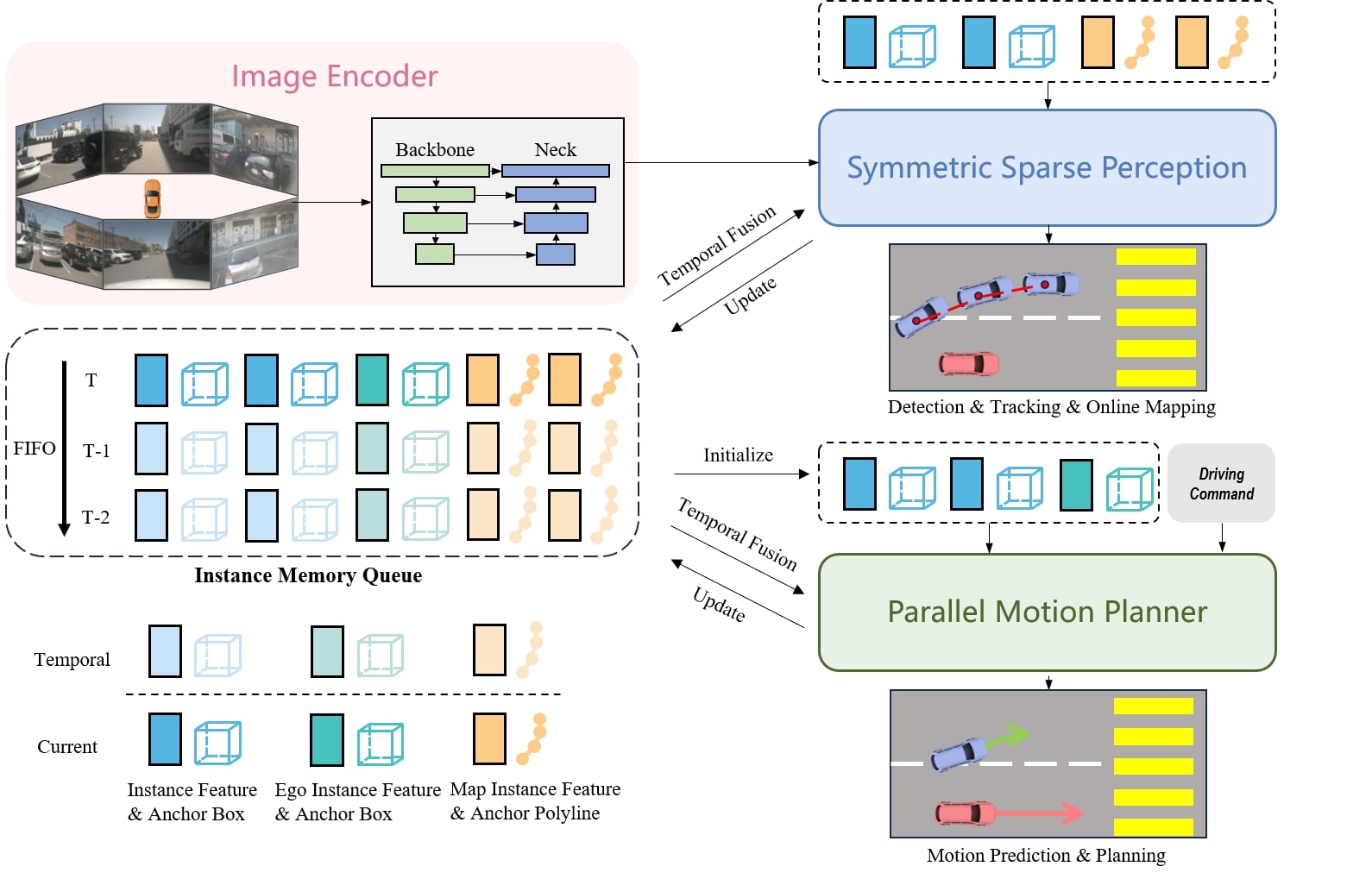}
    \caption{SparseDrive architecture~\cite{sun2024sparsedrive}, featuring sparse query-based perception and parallel motion prediction and planning. We inject EC-IoU into the detection loss while keeping the remaining modules and losses unchanged. Adapted from Fig.~2 in~\cite{sun2024sparsedrive}~\textcopyright~2025~IEEE.}
    \label{fig:e2e:sparsedrive}
\end{figure}

\subsection{SparseDrive and EC-IoU Integration}
\label{sec:e2e:method}

We adopt SparseDrive~\cite{sun2024sparsedrive}, a state-of-the-art perception-to-planning model that combines sparse, query-based 3D perception with parallel motion prediction and planning. Compared with earlier E2E stacks such as UniAD~\cite{hu2023uniad}, SparseDrive avoids dense BEV feature computation, achieves near-real-time inference, and substantially reduces collision rates in simulation. In its original design, SparseDrive is trained with the multi-task objective
\begin{equation}
\label{eq:e2e:overall_loss}
L \defeq L_{\mathrm{det}} + L_{\mathrm{map}} + L_{\mathrm{motion}} + L_{\mathrm{plan}} + L_{\mathrm{depth}},
\end{equation}
where $L_{\mathrm{det}}$ supervises 3D object detection, and the remaining terms supervise map elements, motion forecasting, ego planning, and depth, respectively~\cite{sun2024sparsedrive}. Fig.~\ref{fig:e2e:sparsedrive} shows the corresponding architecture.

Originally, SparseDrive optimizes detection mainly for accuracy using an $\ell_1$-style regression loss. To harden perception in a safety-aware manner, we augment the detection objective with the proposed EC-IoU loss:
\begin{equation}
\label{eq:e2e:det_loss}
L_{\mathrm{det}} \defeq L_{\ell_1} + L_{\mathrm{EC\text{-}IoU}\text{-}\alpha},
\end{equation}
where $\alpha$ controls the ego-centric weighting strength. We evaluate $\alpha \in \{2,4\}$ and keep all other loss terms and training hyperparameters unchanged to isolate the effect of the safety-aware perception objective.

\subsection{Experimental Setup}
\label{sec:e2e:setup}

We train and evaluate the variants on nuScenes using the SparseDrive evaluation protocol~\cite{caesar2020nuscenes,sun2024sparsedrive}. Performance is examined across four tasks: detection, using NDS together with true-positive IoU and EC-IoU; tracking, using Average Multi-Object Tracking Accuracy (AMOTA); motion prediction, using Average Distance Error (ADE); and motion planning, using collision rate (Col.) and the $\ell_2$ distance to a human-driving reference (L2). Each reported SparseDrive result is averaged over ten training trials. Training is performed on our server with four NVIDIA L40S GPUs, equivalent to the eight NVIDIA RTX 4080 GPUs used in the original implementation in terms of memory size, with the same training hyperparameters. Each training run spans ten epochs on nuScenes and takes approximately four hours.

\vspace{-1mm}
\subsection{Results and Discussion}
\label{sec:e2e:results}

Tab.~\ref{tab:e2e:results} shows that incorporating EC-IoU into the joint end-to-end objective yields a clear system-level safety benefit. Relative to the $\ell_1$ baseline, EC-IoU reduces the collision rate from $0.111\%$ to $0.086\%$ for $\alpha=2$, corresponding to a $22.5\%$ reduction, and further to $0.078\%$ for $\alpha=4$, corresponding to a $29.7\%$ reduction.

At the perception level, EC-IoU improves the safety-oriented EC-IoU measure while largely preserving accuracy-oriented metrics such as NDS and IoU. This indicates that an emphasis on ego-critical coverage in the detection training objective can indeed propagate through joint optimization toward safer planning, and that perception-level safety-oriented indicators remain relevant. 

\begin{table*}[t]
\caption{End-to-end driving performance with standard $\ell_1$ loss or EC-IoU loss applied to SparseDrive fine-tuning. UniAD is included for reference. Performance is evaluated for detection, tracking, prediction, and planning. Lower is better for ADE, Col., and L2.}
\label{tab:e2e:results}
\centering
\footnotesize
\setlength{\tabcolsep}{4.0pt}
\begin{tabular}{lccccccc}
\toprule
\multirow{2}{*}{Model} &
\multicolumn{3}{c}{Detection $\uparrow$} &
Track. $\uparrow$ &
Pred. $\downarrow$ &
\multicolumn{2}{c}{Planning $\downarrow$} \\
\cmidrule(lr){2-4}\cmidrule(lr){7-8}
& NDS (\%) & IoU (\%) & EC-IoU (\%) & AMOTA (\%) & ADE (m) & Col. (\%) & L2 (m) \\
\midrule
UniAD~\cite{hu2023uniad}               & 49.80 & -- & -- & 35.90 & 0.71 & 0.610 & 0.73 \\
SparseDrive~\cite{sun2024sparsedrive}  & \textbf{52.43} & \textbf{43.97} & 42.46 & \textbf{36.97} & \textbf{0.62} & 0.111 & \textbf{0.59} \\
\quad w/ EC-IoU-2                      & \underline{51.68} & \underline{43.67} & \underline{43.15} & \underline{36.47} & \underline{0.63} & \underline{0.086} & \underline{0.60} \\
\quad w/ EC-IoU-4                      & 51.46 & 43.51 & \textbf{43.25} & 36.23 & \underline{0.63} & \textbf{0.078} & \underline{0.60} \\
\bottomrule
\end{tabular}
\end{table*}

\begin{figure}[]
    \centering
    \includegraphics[width=0.98\linewidth]{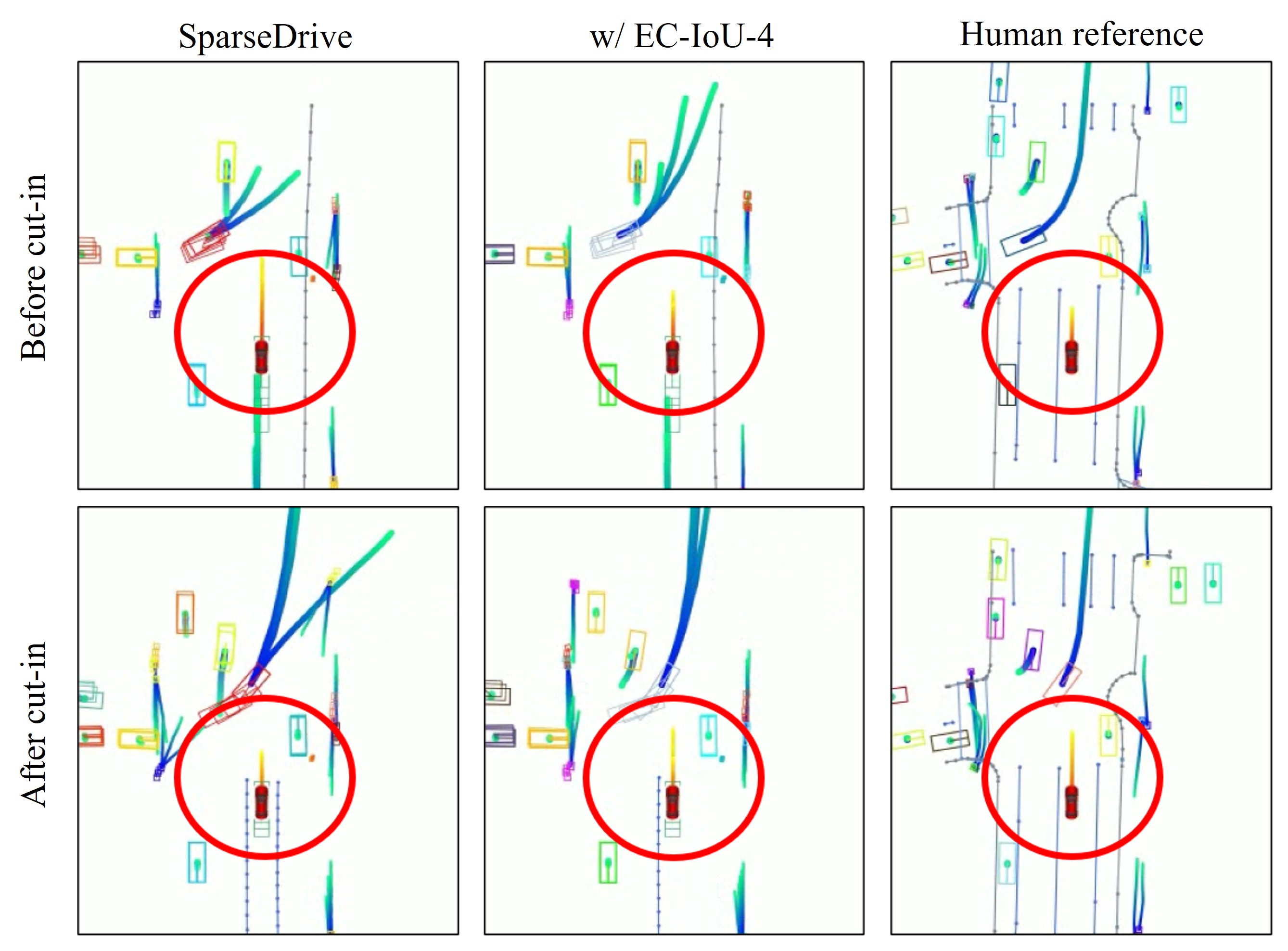}
    \caption{Qualitative comparison of baseline and EC-IoU-optimized SparseDrive variants. In the same cut-in scenario as Fig.~\ref{fig:single:nuscenes_eciou}, the EC-IoU-based model decelerates earlier and accelerates more decisively, more closely matching the human reference.}
    \vspace{-1mm}
    \label{fig:e2e:cutin}
\end{figure}

Fig.~\ref{fig:e2e:cutin} visualizes a representative cut-in scenario (the same one shown in Fig.~\ref{fig:single:nuscenes_eciou}). Compared with the baseline, the EC-IoU-based model slows down earlier before the cut-in and re-accelerates more decisively afterward, generally exhibiting a behavior closer to the human-driving reference.

\vspace{-1mm}
\subsection{Takeaways}
\label{sec:e2e:takeaways}

\begin{itemize}
    \item \textbf{Safety-aware perception hardening transfers to end-to-end driving:} incorporating EC-IoU into the detection training objective reduces collision rate by nearly $30\%$.
    \item \textbf{System-level gains are consistent with improved safety-oriented perception score:} Alongside the reduced collision rate, fine-tuning with EC-IoU also leads to an improved safety-oriented detection score, suggesting the relevance of the perception-level measure.
\end{itemize}

\vspace{-2mm}
\section{Conclusion}
\label{sec:conclusion}

This paper studied how to \emph{align 3D object detection with driving safety} by emphasizing perception errors that are disproportionately consequential to downstream decision making. Building on our prior work on safety-oriented evaluation and safety-aware fine-tuning, we presented three extensions covering single-vehicle perception, AV--infrastructure cooperative perception, and end-to-end driving perspectives.

First, an expanded single-vehicle study across diverse architectures and sensing modalities showed that improvements under conventional benchmarks (e.g., mAP and NDS) do not necessarily translate to safety-oriented gains and safety-aware fine-tuning consistently improved safety-critical localization as well as standard accuracy. Second, we extended safety-oriented assessment to AV--infrastructure cooperative perception by evaluating cooperative models from the ego-vehicle perspective and with safety-aware matching, confirming that cooperation improves ego-relevant perception while revealing ego-centric localization biases that are not emphasized by standard IoU-based evaluation. Third, we demonstrated that safety-aware perception hardening transfers beyond modular pipelines: injecting EC-IoU into a end-to-end perception-to-planning model reduced collision rate and improved system-level safety.

Several directions are promising for future work. The first direction is to incorporate additional sources of safety criticality beyond ego-centric distance, such as time-to-collision, road geometry, intent, and interaction context. Another direction is to develop principled safety impact evaluation for cooperative perception, including closed-loop simulation and causal analyses that link perception improvements to collision reduction at intersections. Third, evaluation and optimization should be generalized to a broader range of scenarios, with particular attention to reliability under distribution shifts and long-tail conditions. Overall, the results support safety-aligned perception as a practical and scalable path toward safer autonomy.

\bibliographystyle{IEEEtran}
\bibliography{references.bib}

\end{document}